\documentclass[letterpaper, 10 pt, conference]{ieeeconf}  

\IEEEoverridecommandlockouts                              

\usepackage{graphics} 
\usepackage{graphicx}
\usepackage{amsmath} 
\usepackage{xcolor}
\usepackage{multirow}
\usepackage{bm}
\usepackage{makecell}
\usepackage{url}
\usepackage{float}
\usepackage{subcaption}
\usepackage{booktabs,makecell,multirow}
\usepackage{amsfonts}
\usepackage{mathrsfs}
\usepackage{cite}
\usepackage{color}
\usepackage{amsmath,amsfonts,amssymb}
\usepackage{amsfonts}
\usepackage[some]{background}
\SetBgScale{1}
\SetBgContents{\parbox{10cm}{%
  \Huge Draft:  \today\\[18cm]\rotatebox{180}{\Huge Draft:  \today}}}
\SetBgColor{red}
\SetBgAngle{270}
\SetBgOpacity{0.2}

\title{\LARGE \bf
Fixed-time Adaptive Neural Control for Physical Human-Robot Collaboration with Time-Varying Workspace Constraints}
\author{Yuzhu Sun, Mien Van*, Stephen McIlvanna, Nguyen Minh Nhat, Seán McLoone\\
	Dariusz Ceglarek and Shuzhi Sam Ge 
\thanks{Yuzhu Sun, Stephen McIlvanna, Nguyen Minh Nhat, Seán McLoone and  Mien Van (corresponding author) are with the Centre for Intelligent Autonomous Manufacturing Systems, School of Electronics, Electrical Engineering and Computer Science, Queen's University Belfast, Northern Ireland, UK. 
        (email: {ysun32, smcilvanna01, nnhat01, s.mcloone, m.van}@qub.ac.uk)}%
\thanks{Dariusz Ceglarek is with the Warwick Manufacturing Group, University of Warwick, Coventry, UK
	(email:  D.J.Ceglarek@warwick.ac.uk)}%
\thanks{Shuzhi Sam Ge is with the Department of Electrical and Computer Engineering and Social Robotics Laboratory, National University of Singapore, Singapore
	(email:  samge@nus.edu.sg)}%
}

\begin{document}

\onecolumn 
\pagestyle{empty} 
\begin{center}
	\large\bfseries  
	This work has been submitted to the IEEE for possible publication. Copyright may be transferred without notice, after which this version may no longer be accessible
\end{center}
\twocolumn 
\setcounter{page}{1} 

\maketitle
\thispagestyle{empty}
\pagestyle{empty}


\begin{abstract}
Physical human-robot collaboration (pHRC) requires both compliance and safety guarantees since robots coordinate with human actions in a shared workspace. This paper presents a novel fixed-time adaptive neural control methodology for handling time-varying workspace constraints that occur in physical human-robot collaboration while also guaranteeing compliance during intended force interactions. The proposed methodology combines the benefits of compliance control, time-varying integral barrier Lyapunov function (TVIBLF) and fixed-time techniques, which not only achieve compliance during physical contact with human operators but also guarantee time-varying workspace constraints and fast tracking error convergence without any restriction on the initial conditions. Furthermore, a neural adaptive control law is designed to compensate for the unknown dynamics and disturbances of the robot manipulator such that the proposed control framework is overall fixed-time converged and capable of online learning without any prior knowledge of robot dynamics and disturbances. The proposed approach is finally validated on a simulated two-link robot manipulator. Simulation results show that the proposed controller is superior in the sense of both tracking error and convergence time compared with the existing barrier Lyapunov functions based controllers, while simultaneously guaranteeing compliance and safety.

\end{abstract}

\section{INTRODUCTION}\label{sec1}

The past few decades have seen rapid development in robot technology and its applications, which allows humans and robots to execute a variety of complex tasks in a shared workspace \cite{RN111}.
To guarantee safety during tasks, robots and human operators have been organised in completely separate areas.  
However, physical contact between humans and robots is unavoidable in some specific scenarios, such as rehabilitation robots \cite{RN100} which guide a patient's arm while coordinating with human movements with a natural fluidity, and collaborative industrial robots performing shared tasks such as holding or co-carrying a load with human partners. Three nested levels of safe human-robot collaboration are given in \cite{RN633}, namely: (i) \emph{safety}, which is realized by detecting and isolating any unintended collisions in the presence of human operators; (ii) \emph{coexistence}, which allows humans and robots to work in the same workspace without the coordination of actions, and; (iii) \emph{collaboration}, which includes physical and contactless collaboration. Among these, physical collaboration requires the robot to be capable of coordinating with human motions with intended physical contacts and exchanging forces with the human in a safe way. Physical collaborative robots make full use of the reasoning capabilities of human partners, and the high precision, repeatability and heavy-duty task execution capabilities of robots \cite{RN634}\cite{RN631}, enabling them to perform much more complicated tasks compared to traditional automated robotic systems \cite{RN546}. During tasks, \emph{coexistence} is the primary mode of operation, and \emph{collaboration} occurs when monitoring signals generated from sensors (e.g., force/torque sensors or camera) detect the presence of physical human contacts. This raises the question of how safe physical human-robot collaboration (pHRC) can be conducted during \emph{coexistence} whilst ensuring \emph{compliance} during \emph{collaboration}.

\begin{figure}[h]
	\centering
	\includegraphics[width=0.48\textwidth]{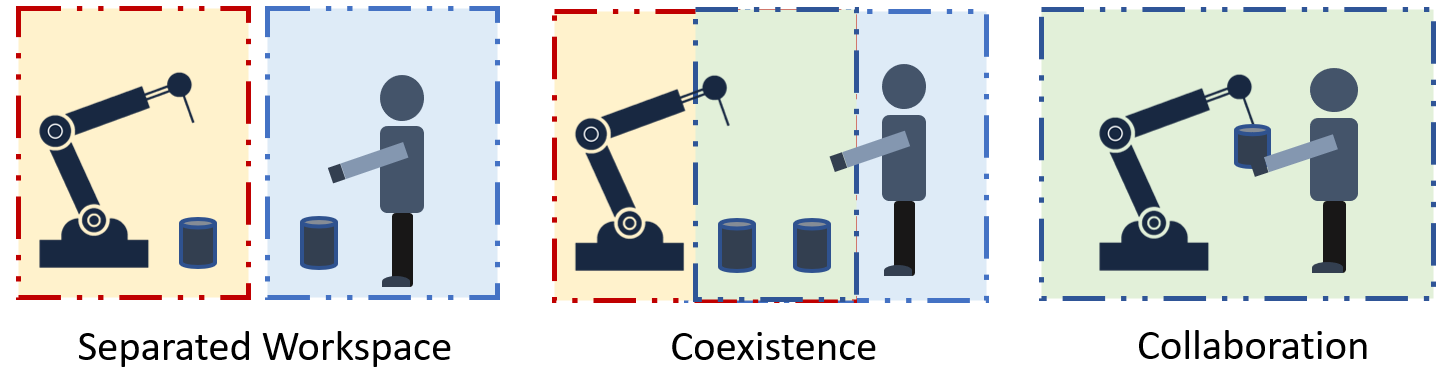}
	\caption{Physical human-robot collaboration.}
	\label{ThreeLayer}
\end{figure}

Traditional position control, whose task is to follow a specific trajectory as accurately as possible, rejects the interaction forces between the robot and human as disturbances \cite{RN100}. To achieve so-called \emph{compliance} during physical contact, there is a vast body of work on impedance/admittance based compliance control in robotics \cite{RN636,RN343,RN103,RN101,RN539}. Compliance control considers both position and force in order to obtain robot movements that are smoother, softer and more human-friendly in the presence of external human forces. Therefore, in this work, we apply admittance control during \emph{collaboration}. To be more specific, we derive the reference trajectory based on the desired task trajectory and force feedback from force/torque sensors. By following such a reference trajectory rather than the original one, robots can coordinate with human motions and comply with human forces during the intended physical contact.

Another major factor with collaborative robots that hinders their use in the real world is safety \cite{RN628}. Robots are complex and fragile. The risk of unwanted collisions between the robot and the surrounding environment (e.g., workspace boundaries, moving obstacles, etc.) exists during production, setup, and regular maintenance. Unintended physical contact can hurt humans with serious consequences. This raises the question of how safety can be strictly assured during the \emph{coexistence} process. Barrier Lyapunov functions (BLFs) based constraint control \cite{RN110,RN102,RN104,RN105}, which enforces safety from the control perspective, has been one of the most effective tools for dealing with constraint problems in control systems \cite{RN106}. BLFs can be generally divided into logarithmic BLFs \cite{RN107}\cite{RN102}, tangent BLFs \cite{RN108}\cite{RN109}, and integral BLFs \cite{RN594,RN587,RN103,RN581,RN583}. Compared with other types of BLFs, integral BLF (IBLF) can directly restrict system states within a certain range without transforming state constraints into error constraints \cite{RN588}\cite{RN582}. The design process using IBLF can therefore be greatly simplified and relaxed in terms of feasibility conditions \cite{RN106}. Subsequently, time-varying IBLF (TVIBLF) \cite{RN588}\cite{RN582} was developed to handle time-varying constraints which are more common in many practical engineering systems since safety boundaries in the workspace are usually time-varying (e.g., when dealing with moving obstacles or a human operator) \cite{RN582}. In \cite{RN588}, TVIBLF combined with the backstepping technique is first introduced for adaptive control of nonlinear systems. In \cite{RN582}, TVIBLF combined with fuzzy logic systems is introduced for a class of strict-feedback nonlinear systems. 

In addition to achieving safe operation, the designer expects the control system to meet the required performance in the shortest possible time \cite{RN590}. Compared with existing finite-time control results \cite{RN590,RN639,RN640,RN375}, the convergence time of a fixed-time controller \cite{RN592} can get rid of the influence of initial conditions and be pre-designed based on parameters of the controller. Studies have shown that fixed-time convergence can produce better tracking performance and robustness to disturbances \cite{RN260}\cite{RN414}. Despite many advantages, only a few studies have integrated fixed-time techniques into BLFs-based constrained control \cite{RN625}\cite{RN270} for better tracking performance, which is essential because better tracking performance with constraints means higher working safety and efficiency of the physical human-robot collaboration. In \cite{RN625}, a novel fixed-time convergent time-varying BLFs-based control scheme is proposed for uncertain nonlinear systems. In \cite{RN270}, a nonsingular adaptive fixed-time switching control method for a class of strict-feedback nonlinear systems is proposed. So far there is little literature that integrates fixed-time techniques into TVIBLF to improve the performance and robustness of the system. Meanwhile, model-free control is a common approach within robotics control literature, since the performance of model-based controllers is dependent on the accuracy of the model. However, even for fully rigid robots, we still need to consider the uncertainties and disturbances that are not modelled such as motor/actuator errors, unintended external human forces and the influence of the noisy environment \cite{RN261}. Apart from this, model-free controllers can be compatible with robotic systems which have different dynamic models, and therefore be more practical in real-world scenarios. To suppress these problems, fuzzy logic systems (FLSs) \cite{RN625}\cite{RN232} and neural networks (NNs)\cite{RN584,RN585,RN192} have long been introduced to estimate uncertainties inherent in practical systems and the influences of unknown dynamics. 

\begin{figure}[t]
	\centering
	\includegraphics[width=0.48\textwidth]{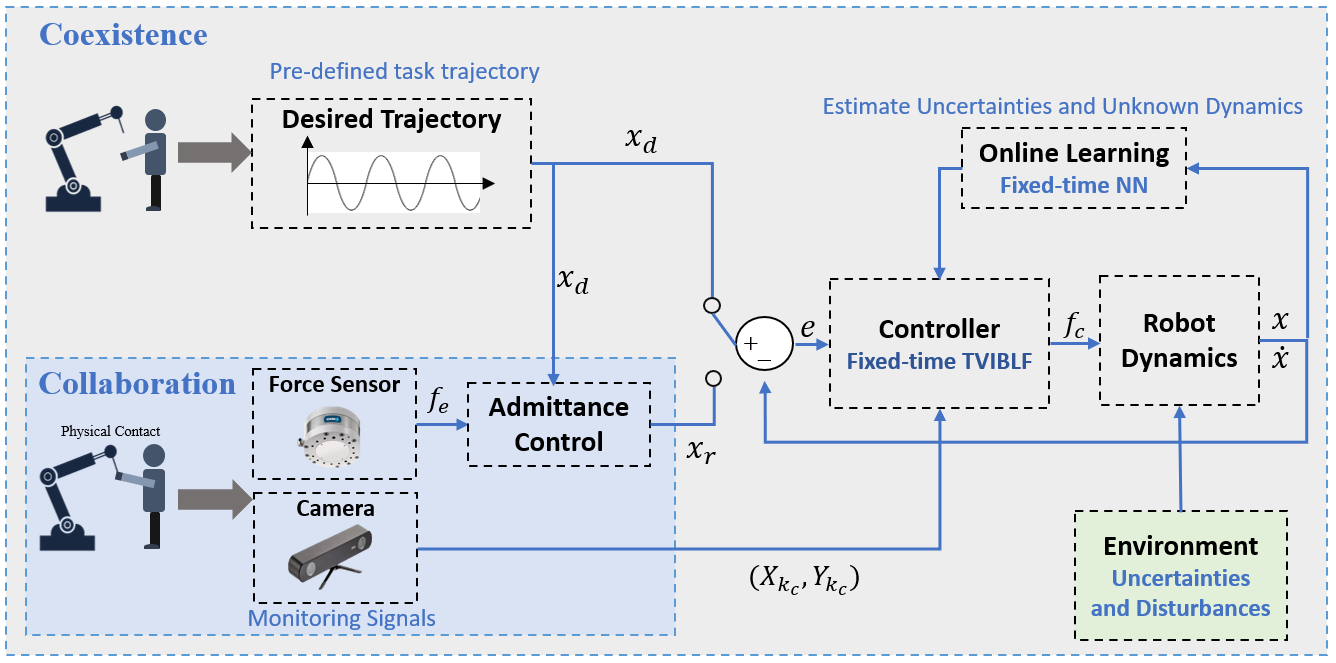}
	\caption{The structure of proposed control framework.}
	\label{Structure}
\end{figure}

Motivated by the above discussion, in this paper, a novel fixed-time control technique that integrates the fixed-time technique into the time-varying integral barrier Lyapunov function during the backstepping control design process. To eliminate the impact of unknown dynamics and model uncertainties, a novel neural network adaptive law is designed to achieve overall fixed-time convergence of the system, which further emphasises the novel contribution of this paper. The overall structure of the proposed control framework is shown in Fig. \ref{Structure}. The contributions and innovations of the proposed approach can be highlighted in a comparison with other approaches as follows:
\begin{enumerate}
	\item Compared with existing IBLF and TVIBLF based constrained control \cite{RN103}\cite{RN588}, the proposed controller (FxTTVIBLF) integrates the fixed-time technique into the backstepping control design process to derive a controller with fixed-time convergence. Such a controller provides better tracking performance with lower tracking error, fast convergence and higher robustness to the disturbances without any dependency on initial conditions. Meanwhile, a neural network adaptor is designed to approximate the unknown dynamics and uncertainties online such that the proposed control framework is overall fixed-time convergent and compatible with different robotic systems.
	\item Compared with existing control methods for physical human-robot collaboration, the proposed approach guarantees the time-varying safety constraints with better tracking performance during coexistence, while simultaneously guaranteeing compliance when physical collaboration occurs. Such a controller is more practical in real-world scenarios since safety and compliance are two essential concerns that need to be addressed during physical collaboration.
\end{enumerate}

The remainder of this paper is organized as follows. The general mathematical model of the robot manipulator, admittance control, neural networks and the problem formulation are presented in Section \ref{sec2}.  The design process of fixed-time time-varying constrained control and neural network adaptor are developed in Section \ref{sec3}. Simulation results of the proposed system are presented in Section \ref{sec4}. Finally, Section \ref{sec5} discusses the conclusions and future work.

\section{Problem Formulations and Preliminaries}\label{sec2}
In this section, we begin by briefly introducing the dynamic model of the robot manipulator, the basics of admittance control, neural networks and overall problem formulation. In addition, some important lemmas are also given in this section, which pave the way for the control design and the proof of stability.
\subsection{Robot Dynamic Model}
The dynamic model of the robot describes the relationship between force and motion. In joint space, the dynamics of a robot manipulator can be written as:
\begin{equation}
	\label{eq1}
	M\left( q \right) \ddot{q}+C\left( q,\dot{q} \right) \dot{q}+G\left( q \right) +F\left(q, \dot{q} \right) =\tau _c +\tau _e
\end{equation}
where $q=\left[ q_1,q_2,...,q_n \right]^T$ is the vector of joint angles, $n$ is the number of the degree of freedom (DOF) of the robot manipulator, and $\dot{q}$ and $\ddot{q}$ are the joint velocities and accelerations, respectively. $M\left( q \right)$ is the mass matrix, $C\left( q,\dot{q} \right)$ is the Coriolis and centrifugal forces matrix, $G\left( q \right)$ is the gravity matrix and $F\left(q, \dot{q} \right)$ is the friction matrix for the manipulator. The $M\left( q \right)$, $C\left( q,\dot{q} \right)$ and $G\left( q \right)$ terms contain uncertainties and $F\left(q, \dot{q} \right)$ represent disturbances. $\tau_c$ is the control torque generated by the controller that we are going to design in the following sections, and $\tau_e$ is the external torque from the human operator. Employing the joint space dynamics of the robot can simplify the mathematics of the relationship between each joint of the robot. However, in real-life scenarios, the task trajectory and safety constraints for obstacles are always described in the Cartesian space. The transformation between joint angle velocities and Cartesian velocities of the robot manipulator can be written as:
\begin{equation}
	\label{eq2}
	\dot{x}=J\left( q \right) \dot{q}
\end{equation}
where $J\left( q \right)$ is the Jacobian of the robot manipulator. To simplify the problem, we assume the Jacobian is known and non-singular in this paper. Using (\ref{eq2}), we can transfer the joint space dynamics of the robot (\ref{eq1}) into Cartesian space as:
\begin{equation}
	\label{eq3}
	M_x\ddot{x}+C_x\dot{x}+G_x+F_x=f_c+f_e
\end{equation}
where $x=\left[ x_1,x_2,...,x_m \right]^T$ is the position of the robot end-effector in Cartesian space. To simplify the problem, we assume the robot is non-redundant ($m=n$). $\dot{x}$ and $\ddot{x}$ are the Cartesian velocity and acceleration. $f_c=J^{-T}\left( q \right) \tau _c$, and $f_e=J^{-T}\left( q \right) \tau _e$ denote the control forces and the external human forces, respectively. The coefficient matrices transferred to Cartesian space are given as:

\begin{small}
	\begin{equation}
		\label{eq4}
		\begin{array}{c}
			M_x=J^{-T}\left( q \right) M\left( q \right) J^{-1}\left( q \right)\\
			C_x=J^{-T}\left( q \right) \left( C\left( q,\dot{q} \right) -M\left( q \right) J^{-1}\left( q \right) \dot{J}\left( q \right) \right) J^{-1}\left( q \right)\\
			G_x=J^{-T}\left( q \right) G\left( q \right) , F_x=J^{-T}\left( q \right) F\left(q, \dot{q} \right)\\
		\end{array}
	\end{equation}
\end{small}

The following important properties pertaining to the robot dynamic equations can be exploited to good advantage for control design \cite{RN222}:

\emph{Property 1}: The matrix $M_x$ is symmetric positive definite.

\emph{Property 2}: The matrix $\dot{M}_x-2C_x$ is skew symmetric.

\subsection{Admittance Control}
\begin{figure}[h]
	\centering
	\includegraphics[width=0.2\textwidth]{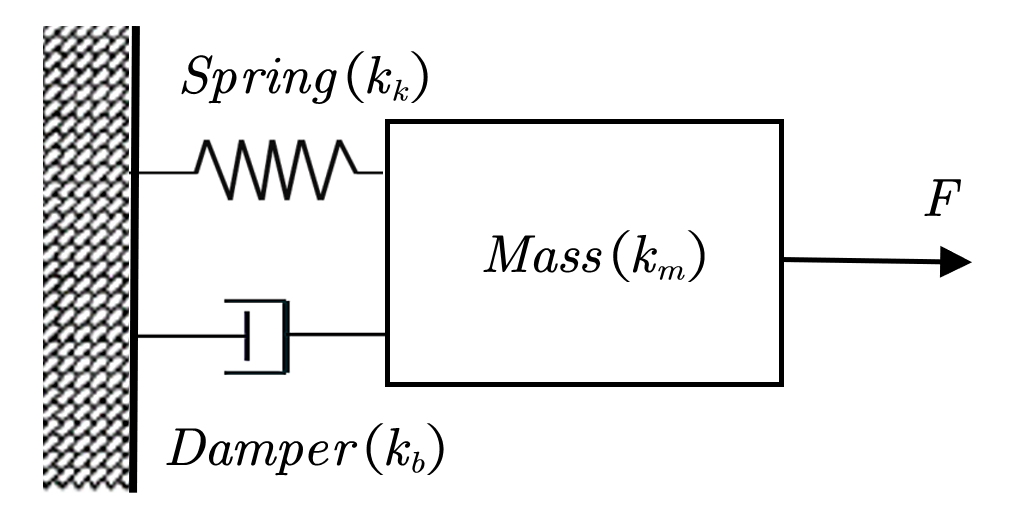}
	\caption{Mass-spring-damper system.}
	\label{Mass-spring-damper}
\end{figure}

To provide compliance for physical human-robot force interaction, the contact point between the human and the robot is modelled as a mass-spring-damper system to imitate human muscle mechanisms. The virtual mass, spring, and damper ensure that the interaction forces are elastic and never vibrate at the contact point, as depicted in Fig. \ref{Mass-spring-damper}. 

Impedance and admittance are two opposite notions in compliance control. The system is regarded as admittance when the input is force and the output is position, while it is impedance when the input is position and the output is force. In this work, we are going to derive the trajectory which performs compliance behaviour based on the predefined trajectory and force feedback from force sensors. Therefore, we apply admittance control in our proposed framework. Since we need to implement Cartesian compliance, we assign the stiffness and damping at the end-effector level as follows:
\begin{small}
	\begin{equation}
		\label{eq10}
		k_{m_i}\left( \ddot{x}_{r_i}-\ddot{x}_{d_i} \right) +k_{b_i}\left( \dot{x}_{r_i}-\dot{x}_{d_i} \right) +k_{k_i}\left( x_{r_i}-x_{d_i} \right) =f_{e_i}
	\end{equation}
\end{small}

\noindent Here $k_{m_i}$, $k_{b_i}$ and $k_{k_i}$ are the mass, spring and damping coefficients for each dimension, $i=1,2,...,m$. $x_{d_i}$ is the desired trajectory which is pre-defined to finish the task, and $x_{r_i}$ is the reference trajectory which we pursue to perform compliance behaviours in response to external human forces $f_{e_i}$. When we define the $x_{d_i}$ and detect the human force $f_{e_i}$ via force sensors, $x_{r_i}$ can be derived by integrating (\ref{eq10}) twice.

\subsection{Radial Basis Functions Neural Networks}
Radial Basis Functions Neural Networks (RBFNNs), commonly used for function approximation problems, contain three layers: the input layer, hidden layer and output layer. The input layer consists of predictor variables $X=\left[ x_1,x_2,...,x_r \right]^T$ which are considered to be useful or informative with respect to the output. $r$ is the number of the input variables. The hidden layer contains a variable number of neurons $\varPhi\left( X \right) =\left[ \varPhi_1\left( X \right) ,\varPhi_2\left( X \right) ,...,\varPhi_l\left( X \right) \right]^T$, where $l$ is the number of hidden nodes. Each neuron comprises a Gaussian radial basis function which is defined as:
\begin{equation}
	\label{RBFNN1}
	\varPhi_i\left( X \right) =\exp \left( \frac{-\left( X-C_i \right) ^T\left( X-C_i \right)}{B_{i}^{2}} \right) , i=1,2,...,l
\end{equation}
where $C_i=\left[ c_{i1}c_{i2},...,c_{ir} \right]^T $ is the centre and $B_i$ is the width of $i$-th Gaussian radial basis functions. The output of the RBFNN is given by:
\begin{equation}
	\label{RBFNN2}
	\hat{H}\left( X \right) =\hat{\theta}^{T}\varPhi\left( X \right)
\end{equation}
Here, $\hat{\theta}=\left[ \hat{\theta}_1,\hat{\theta}_2,...,\hat{\theta}_l \right] ^T$ denotes the estimation of the optimal weights $\theta$ associated with the hidden nodes. The RBFNN in (\ref{RBFNN2}) can approximate functions to arbitrary accuracy when the number of hidden nodes is large enough. Therefore, the optimal output $H\left( X \right)$ can be expressed as
\begin{equation}
	\label{RBFNN2_1}
	H\left( X \right) =\theta^{T}\varPhi\left( X \right) + \epsilon
\end{equation}
where $\epsilon$ denotes the approximation error of the RBFNN. The optimal weight $\theta$ is obtained by minimizing the estimation error over the training set. 

\begin{equation}
	\label{RBFNN3}
	\theta =arg\underset{\theta \in \mathbb{R} ^l}{\min}\left\{ \mathop {sup} \limits_{X\in \varOmega _X}\left| H\left( X \right) -\theta ^T\varPhi \left( X \right) \right| \right\} 
\end{equation}

\emph{Assumption 1}\cite{RN584}: The approximation error $\epsilon$ is bounded by $\left| \epsilon \right|\le \bar{\epsilon}$ with the constant $\bar{\epsilon}>0$.

\subsection{Problem Formulation}
For ease of the backstepping control design, let $\eta _1=x, \eta _2=\dot{x}$ and $u=f_c$, Cartesian dynamics (\ref{eq3}) can then be written as:
\begin{equation}
	\label{eq5}
	\begin{cases}
		\dot{\eta}_1=\eta _2\\
		\dot{\eta}_2={M_x}^{-1}\left( -C_x\eta_2-G_x-F_x+f_e+u \right)\\
	\end{cases}
\end{equation}

The control object of this paper is to make task space variable $x$ track a desired trajectory $x_d$ whilst complying with human force $f_e$. In addition, $x$ is constrained by the time-varying workspace safety constraints $k_c\left( t \right) =\left[ k_{c_1}\left( t \right) ,k_{c_2}\left( t \right) ,...,k_{c_n}\left( t \right) \right]^T $ to satisfy $\left|x_i\right|\le k_{c_i}\left(t\right)$, where $i=1,2,...,n$. The following lemmas will be useful in the control design in the next section.

\noindent\textbf{\emph{Lemma 1}} \cite{RN773}: 
Consider a nonlinear system:
\begin{equation}
	\label{eq7}
	\dot{x}=f\left( x,t \right), x\left(0\right)=x_0
\end{equation}
where $x\in R^n$, and $f:\ R^n\rightarrow R^n$ is a nonlinear function. If there exists a Lyapunov function $V\left(x\right)$ such that:
\begin{equation}
	\label{eq8}
	\dot{V}\left( x \right) \le -\alpha V^{p_c}\left( x \right) -\beta V^{q_c}\left( x \right) + \sigma
\end{equation}
where $\alpha ,\beta >0, p_c>1, 0<{q_c}<1$ and $\sigma$ is a small positive constant, then the system (\ref{eq7}) is practical fixed-time stable and the residual set of the system solution is given by:

\begin{small}
\begin{equation}
	\varOmega _x=\left\{ x|V\left( x \right) \le \min \left\{ \left[ \frac{\sigma}{\alpha \left( 1-v \right)} \right] ^{\frac{1}{p_c}},\left[ \frac{\sigma}{\beta \left( 1-v \right)} \right] ^{\frac{1}{{q_c}}} \right\} \right\} 
\end{equation}
\end{small}

\noindent where $v$ is a scalar and satisfies $0<v\le1$. Then, the time $T$ which is needed to reach the residual set is bounded by:

\begin{equation}
	\label{Lemma1}
	T\le T_{\max}:=\frac{1}{\alpha v \left( p_c-1 \right)}+\frac{1}{\beta v\left( 1-{q_c} \right)} 
\end{equation}

\noindent\textbf{\emph{Lemma 2}}\cite{RN587}: Consider an Integral Barrier Lyapunov Functions (IBLFs) candidate:
\begin{equation}
	\label{eq8_2}
	V_i=\int\limits_0^{z_{1_i}}{\frac{\delta k_{c_i}^{2}}{k_{c_i}^{2}-\left( \delta +x_{d_i} \right) ^2}d\delta}
\end{equation}

The function described in (\ref{eq8_2}) satisfies the following property for any $\left|\eta _{1_i}\right|<k_{c_i}$:
\begin{equation}
	\label{Lemma2}
	V_i\le \frac{k_{c_i}^{2}z_{1_i}^{2}}{k_{c_i}^{2}-\eta _{1_i}^{2}}
\end{equation}

\noindent\textbf{\emph{Lemma 3}} \cite{RN643}: Let $\tilde{W}=W-\hat{W}$, for any $0<{q_c} <1$, where ${q_c} =\frac{{q_c} _1}{{q_c} _2}$ and ${q_c} _1$, ${q_c} _2$ are odd integers, the following inequality holds:
\begin{equation}
	\label{Lemma3}
	\tilde{W}\hat{W}^{{q_c}}\le n_1W^{{q_c} + 1}-n_2\tilde{W}^{{q_c} +1}
\end{equation}
where $n_1=\left( \frac{1}{1+{q_c}} \right) \left( 1-2^{{q_c} -1}+\frac{{q_c}}{1+{q_c}}+\frac{2^{{q_c}}\left( 1-{q_c} ^2 \right)}{1+{q_c}} \right)$ and $n_2=\frac{2^{{q_c}}-1}{1+{q_c}}\left( 1-2^{{q_c}\left( {q_c}-1 \right)} \right)$.

\noindent\textbf{\emph{Lemma 4}} \cite{RN443}: For $x_i\ge 0$, $p_c>1$ and $0<{q_c}<1$, the following inequalities hold: 

\begin{equation}
	\begin{aligned}
	\label{Lemma4}
	n^{1-p_c}\left( \sum_{i=1}^n{x_i} \right) ^{p_c}&\le \sum_{i=1}^n{x_{i}^{p_c}}\\
	\left( \sum_{i=1}^n{x_i} \right) ^{q_c}&\le \sum_{i=1}^n{{x_i}^{q_c}}
\end{aligned}
\end{equation}

\noindent\textbf{\emph{Lemma 5}} \cite{RN644}: Let $a>0$ be a constant and $b<a$, $p_c$ is an odd integer and $p_c>1$, the following inequality holds:

\begin{equation}
	\label{Lemma5}
	b\left( a-b \right) ^{p_c}\le a^{p_c +1}-b^{p_c +1}
\end{equation}

\noindent\textbf{\emph{Lemma 6}} \cite{RN644}: Consider a differential equation of the form:

\begin{equation}
	\label{Lemma6}
	\dot{x}\left( t \right) =-c _1x^{2\mu -1}\left( t \right) -c _2x^{2v-1}\left( t \right) +\sigma \left( t \right) 
\end{equation}
where $x\left( t \right) \in R$, $c_1$ and $c_2$ are positive constants. $\mu =\frac{p_1}{q_1}>1, 1>v=\frac{p_2}{q_2}>\frac{1}{2}$, where $p_1$, $p_2$, $q_1$, $q_2$ are positive odd numbers and $\sigma \left( t \right) $ is a positive function. If $x\left(t_0\right)\ge0$, $x\left(t\right)\ge0$ holds for $\forall t\ge t_0$.

\noindent\textbf{\emph{Lemma 7 (Young's inequality)}} \cite{RN645}: For $\forall \left( x,y \right) \in R^2$, the following inequality holds:

\begin{equation}
	\label{Lemma7}
	xy\le \frac{\varepsilon ^a}{a}\left| x \right|^a+\frac{1}{b\varepsilon ^b}\left| y \right|^b
\end{equation}
where $\varepsilon >0$, $a,b>1$ and $\left( a-1 \right) \left( b-1 \right) =1$.

\noindent\textbf{\emph{Lemma 8}} \cite{RN584}: For $\forall x\in R$, the following inequality holds:

\begin{equation}
	\label{Lemma8}
	\lambda _{\min}\left( M \right) \left\| x \right\| ^2\le x^TMx\le \lambda _{\max}\left( M \right) \left\| x \right\| ^2
\end{equation}
where $\lambda _{\min}\left( M \right)$ and $\lambda _{\max}\left( M \right)$ represents the minimum and maximum eigenvalues of $M$, respectively. $\left\| \cdot \right\| $ represents the Euclidean norm.

\section{Control Design and Stability Analysis}\label{sec3}
In this section, we first give details of the model-based fixed-time time-varying IBLF controller design and how the backstepping method is incorporated within the design process. Then, for compatibility with different robotic systems and complex environments, we introduce NN models to approximate the uncertainties and unknown dynamics. The proposed control framework is overall fixed-time convergent. The proof of the stability and the analysis of the fixed convergence time are given subsequently. Firstly, according to (\ref{eq5}), we define the error variables as follows:
\begin{equation}
	\label{s3_1}
	\begin{aligned}
		&z_1=\eta _1-x_r\\ 
		&z_2=\eta _2-\alpha 
	\end{aligned} 
\end{equation}
where $\eta _1=\left[ \eta _{1_1},\eta _{1_2},...,\eta _{1_n} \right] ^T$ and $\eta _2=\left[ \eta _{2_1},\eta _{2_2},...,\eta _{2_n} \right] ^T$ are states of the system. $x_r=\left[ x _{r_1},x _{r_2},...,x _{r_n} \right] ^T$ is the reference trajectory generated from admittance control (\ref{eq10}). $\alpha =\left[ \alpha _1,\alpha _2,...,\alpha _n \right] ^T$ is the stabilising function to be designed. To address the constraints on $\eta_1$, we consider the integral barrier Lyapunov candidate as follows:

\begin{equation}
	\label{s3_2}
	\begin{aligned}
		V_1\left( z_1,x_r \right) &=\sum_{i=1}^n{V_{1_i}\left( z_{1_i},x_{r_i} \right)}\\ 
		{V_{1_i}}_{\left( z_{1_i},x_{r_i},k_{c_i} \right)}&=\int\limits_0^{z_{1_i}}{\frac{\delta k_{c_i}^{2}}{k_{c_i}^{2}-\left( \delta +x_{r_i} \right) ^2}d\delta}
	\end{aligned} 
\end{equation}
where the variable $\delta$ is a member within the integrating range $\left[0,z_{1_i}\right]$. $k_{c_i}$ and $x_{r_i}$ are the $i$-th element of the constraints and reference trajectory, respectively. It can be seen that ${V_1}$ is positive definite and continuously differentiable. If we design a controller $u$ such that $\dot{V}_1\le 0$, ${V_1}$ satisfies the decrescent condition and is bounded. Therefore, it has to be true that $\left|\eta _{1_i}\right| \ne k_{c_i}$, which means $\eta _{1_i}$ remains in the region $\left|\eta _{1_i}\right|<k_{c_i}$ and the system is safe. The time-derivative of $V_1$ is:
\begin{equation}
	\label{s3_3}
	\begin{aligned}
		\dot{V}_1=\sum_{i=1}^n{\frac{z_{1_i}k_{c_i}^{2}}{k_{c_i}^{2}-\eta _{1_i}^{2}}\dot{z}_{1_i}}+\sum_{i=1}^n{\frac{\partial V_{1_i}}{\partial x_{r_i}}\dot{x}_{r_i}}+\sum_{i=1}^n{\frac{\partial V_{1_i}}{\partial k_{c_i}}\dot{k}_{c_i}}
	\end{aligned} 
\end{equation}

According to \cite{RN588}:
\begin{equation}
	\label{s3_4}
	\begin{aligned}
		\sum_{i=1}^n{\frac{\partial V_{1_i}}{\partial x_{r_i}}\dot{x}_{r_i}}&=\sum_{i=1}^n{\dot{x}_{r_i}\int\limits_0^{z_{1_i}}{\frac{\partial}{\partial x_{r_i}}\frac{\delta k_{c_i}^{2}}{k_{c_i}^{2}-\left( \delta +x_{r_i} \right) ^2}d\delta}}\\
		&=\sum_{i=1}^n{\dot{x}_{r_i}z_{1_i}\left( \frac{k_{c_i}^{2}}{k_{c_i}^{2}-\eta _{1_i}^{2}}-\rho _i \right)}
	\end{aligned} 
\end{equation}
where 

\begin{equation}
	\label{s3_5}
	\begin{aligned}
		\rho _i=\frac{k_{c_i}}{2z_{1_i}}\ln \left( \frac{\left( k_{c_i}+\eta _{1_i} \right) \left( k_{c_i}-x_{r_i} \right)}{\left( k_{c_i}-\eta _{1_i} \right) \left( k_{c_i}+x_{r_i} \right)} \right) 
	\end{aligned} 
\end{equation}
Similarly to (\ref{s3_4}):
\begin{equation}
	\label{s3_6}
	\begin{aligned}
		\sum_{i=1}^n{\frac{\partial V_{1_i}}{\partial k_{c_i}}\dot{k}_{c_i}}&=\sum_{i=1}^n{\dot{k}_{c_i}\int\limits_0^{z_{1_i}}{\frac{\partial}{\partial k_{c_i}}\frac{\delta k_{c_i}^{2}}{k_{c_i}^{2}-\left( \delta +x_{r_i} \right) ^2}d\delta}}\\
		&=\sum_{i=1}^n{\dot{k}_{c_i}z_{1_i}\left( \frac{-z_{1_i}k_{c_i}}{k_{c_i}^{2}-\eta _{1_i}^2}+\omega _i \right)}
	\end{aligned} 
\end{equation}
where

\begin{equation}
	\label{s3_7}
	\begin{aligned}
		\omega _i=&-\frac{x_{r_i}k_{c_i}}{k_{c_i}^{2}-\eta _{1_i}^2}+\frac{k_{c_i}}{z_{1_i}}\ln \left( \frac{k_{c_i}^{2}-\eta _{1_i}^2}{k_{c_i}^{2}-x_{r_i}^2} \right)\\
		 &+\frac{x_{r_i}}{2z_{1_i}}\ln \left( \frac{k_{c_i}^{2}-x_{r_i}^2}{k_{c_i}^{2}-\eta _{1_i}^2} \right) 
	\end{aligned} 
\end{equation}
Since $\dot{z}_{1_i}=z_{2_i}+\alpha _i-\dot{x}_{r_i}$, substitute (\ref{s3_4})-(\ref{s3_7}) into (\ref{s3_3}), we have:
\begin{small}
\begin{equation}
	\label{s3_8}
	\begin{aligned}
		\dot{V}_1=&\sum_{i=1}^n{\frac{z_{1_i}z_{2_i}k_{c_i}^{2}}{k_{c_i}^{2}-\eta _{1_i}^2}}+\sum_{i=1}^n{\frac{\alpha _iz_{1_i}k_{c_i}^{2}}{k_{c_i}^{2}-{\eta_{1_i}}^2}}+\sum_{i=1}^n{\frac{-z_{1_i}^{2}k_{c_i}\dot{k}_{c_i}}{k_{c_i}^{2}-\eta _{1_i}^2}}\\
		&-\sum_{i=1}^n{\frac{z_{1_i}\rho _i\dot{x}_{r_i}\left( k_{c_i}^{2}-\eta _{1_i}^2 \right)}{k_{c_i}^{2}-\eta _{1_i}^2}}+\sum_{i=1}^n{\frac{z_{1_i}\omega _i\dot{k}_{c_i}\left( k_{c_i}^{2}-\eta _{1_i}^2 \right)}{k_{c_i}^{2}-\eta _{1_i}^2}}
	\end{aligned} 
\end{equation}
\end{small}
Design the stabilizing function $\alpha_i$ as:
\begin{small}
\begin{equation}
	\label{s3_9}
	\begin{aligned}
		\alpha _i=&\frac{\left( k_{c_i}^{2}-\eta _{1_i}^2 \right) \dot{x}_{r_i}\rho _i}{k_{c_i}^{2}}-\frac{\left( k_{c_i}^{2}-\eta _{1_i}^2 \right) \dot{k}_{c_i}\omega _i}{k_{c_i}^{2}}+\frac{z_{1_i}\dot{k}_{c_i}}{k_{c_i}}\\
		&-\theta _1\frac{z_{1_i}^{2p_c-1}k_{c_i}^{2p_c-2}}{\left( k_{c_i}^{2}-\eta _{1_i}^2 \right) ^{p_c+1}}-\theta _2\frac{z_{1_i}^{2q_c-1}k_{c_i}^{2q_c-2}}{\left( k_{c_i}^{2}-\eta _{1_i}^2 \right) ^{{q_c}+1}}-\kappa_{1_i}z_{1_i}
	\end{aligned} 
\end{equation}
\end{small}

\noindent where $\kappa_1$ is a positive control gain and $\theta_1,\theta_2>0$, $p_c>1$, $0<{q_c}<1$. Substitute (\ref{s3_9}) into (\ref{s3_8}), we have:

\begin{small}
	\begin{equation}
		\label{s3_10}
		\begin{aligned}
			\dot{V}_1=&\sum_{i=1}^n{\frac{z_{1_i}z_{2_i}k_{c_i}^{2}}{k_{c_i}^{2}-{\eta _{1_i}}^2}}-\sum_{i=1}^n{\frac{\kappa_{1_i}z_{1_i}^{2}k_{c_i}^{2}}{k_{c_i}^{2}-\eta _{1_i}^{2}}}\\
			&-\theta_1\sum_{i=1}^n{\left( \frac{z_{1_i}^{2}k_{c_i}^{2}}{k_{c_i}^{2}-{\eta _{1_i}}^2} \right) ^{p_c}}
			-\theta_2\sum_{i=1}^n{\left( \frac{z_{1_i}^{2}k_{c_i}^{2}}{k_{c_i}^{2}-{\eta _{1_i}}^2} \right) ^{q_c}}
		\end{aligned} 
	\end{equation}
\end{small}

\noindent For any $\left|\eta _{1_i}\right|<k_{c_i}$, it is clear that $\sum_{i=1}^n{\frac{\kappa_{1_i}z_{1_i}^{2}k_{c_i}^{2}}{k_{c_i}^{2}-\eta _{1_i}^{2}}}>0$. Therefore, according to Lemma 2, we have:
\begin{equation}
		\label{s3_112}
		\begin{aligned}
			\dot{V}_1\le \sum_{i=1}^n{\frac{z_{1_i}z_{2_i}k_{c_i}^{2}}{k_{c_i}^{2}-{\eta _{1_i}}^2}}-\theta _1\sum_{i=1}^n{\left( V_{1_i} \right) ^{p_c}}-\theta _2\sum_{i=1}^n{\left( V_{1_i} \right) ^{q_c}}
		\end{aligned} 
\end{equation}

\noindent According to Lemma 4, we have:
\begin{equation}
	\label{s3_11}
	\begin{aligned}
		\dot{V}_1\le \sum_{i=1}^n{\frac{z_{1_i}z_{2_i}k_{c_i}^{2}}{k_{c_i}^{2}-{\eta _{1_i}^2}}}-\lambda _1V_{1}^{p_c}-\lambda _2V_{1}^{{q_c}}
	\end{aligned} 
\end{equation}
where 

\begin{equation}
	\label{s3_1111}
	\begin{aligned}
		\lambda _1&=\theta _1n^{1-p_c}
		\\
		\lambda _2&=\theta _2
	\end{aligned} 
\end{equation}
\noindent\textbf{\emph{Remark 1}}: When implementing the stabilizing function $\alpha$, there will be a singularity problem when $z_1=0$. By using L’Hôpital’s rule, we have:
\begin{equation}
	\label{s3_111}
	\begin{aligned}
		\lim_{z_{1_i}\rightarrow 0} \rho _i&=\frac{k_{c_i}^{2}}{k_{c_i}^{2}-x_{r_i}^{2}}
		\\
		\lim_{z_{1_i}\rightarrow 0} \omega _i&=\frac{x_{r_i}^{2}-3x_{r_i}k_{c_i}}{k_{c_i}^{2}-x_{r_i}^{2}}
	\end{aligned} 
\end{equation}

\noindent Then, we design the controller $u$ to provide stability and fixed-time convergence properties to the system:
\begin{equation}
	\label{s3_14}
	\begin{aligned}
		u=& G_x+F_x+M_x\dot{\alpha}+C_x\alpha -f_e -\frac{k_{c}^{2}z_1}{k_{c}^{2}-\eta _{1}^{2}}\\
		&-k_1z_2-\frac{1}{2^{p_c}}k_2{z_2}^{2p_c-1}-\frac{1}{2^{q_c}}k_3{z_2}^{2q_c-1}
	\end{aligned} 
\end{equation}
where $k_1$ is a positive control gain, $k_2$, $k_3$, $p_c$ and ${q_c}$ are fixed-time constants which satisfy $k_2, k_3 >0$, $p_c>1$ and $0<{q_c}<1$. 

\noindent\textbf{\emph{Theorem 1}}: Consider the system (\ref{eq1}) subject to the constraints $ k_c\left(t\right)$ and external disturbances $F\left({q}, \dot{{q}} \right)$. The system is fixed-time stable and the convergence time of the tracking errors is bounded by applying the stabilizing function (\ref{s3_9}) and the controller (\ref{s3_14}).

\noindent\textbf{\emph{Proof}}: Select a Lyapunov function candidate as:
\begin{equation}
	\label{s3_12}
	\begin{aligned}
		V_2=\frac{1}{2}z_{2}^{T}M_xz_2
	\end{aligned} 
\end{equation}

According to Property 1, Property 2, (\ref{eq5}) and (\ref{s3_1}), the time derivative of $V_2$ is:
\begin{equation}
	\label{s3_13}
	\begin{aligned}
		\dot{V}_2&=z_{2}^{T}M_x\dot{z}_2+\frac{1}{2}z_{2}^{T}\dot{M}_xz_2=z_{2}^{T}\left( M_x\dot{z}_2+C_xz_2 \right)\\ &=z_{2}^{T}\left( u+f_e-G_x-F_x-M_x\dot{\alpha}-C_x\alpha \right)  
	\end{aligned} 
\end{equation}

\noindent Inserting the controller (\ref{s3_14}) into (\ref{s3_13}), according to Lemma 4 and Lemma 8, we have:
\begin{small}
\begin{equation}
	\label{s3_15}
	\begin{aligned}
		\dot{V}_2&=z_{2}^{T}\left( -\frac{k_{c}^{2}z_1}{k_{c}^{2}-\eta _{1}^{2}}-k_1z_2-\frac{1}{2^{p_c}}k_2{z_2}^{2p_c-1}-\frac{1}{2^{q_c}}k_3{z_2}^{2q_c-1} \right) \\
		&\le -\sum_{i=1}^n{\frac{z_{1_i}z_{2_i}k_{c_i}^{2}}{k_{c_i}^{2}-\eta _{1_i}^{2}}}-\lambda _{\min}\left( k_2 \right) n^{1-p_c}\left( \frac{1}{2}\left\| z_2 \right\| ^2 \right) ^{p_c}\\
		&\ \ \    -\lambda _{\min}\left( k_3 \right) \left( \frac{1}{2}\left\| z_2 \right\| ^2 \right) ^{q_c}
	\end{aligned} 
\end{equation}
\end{small}

\noindent Therefore, we have:
\begin{small}
	\begin{equation}
		\label{s3_16}
		\begin{aligned}
			\dot{V}_2\le &- \sum_{i=1}^n{\frac{z_{1_i}z_{2_i}k_{c_i}^{2}}{k_{c_i}^{2}-{\eta _{1_i}}^2}}-\frac{\lambda _{\min}\left( k_2 \right)}{\lambda _{\max}\left( M \right) ^{p_c}} n^{1-p_c}\left( \frac{1}{2}z_{2}^{T}Mz_2 \right) ^{p_c}\\
			&-\frac{\lambda _{\min}\left( k_3 \right)}{\lambda _{\max}\left( M \right) ^{q_c}}\left( \frac{1}{2}z_{2}^{T}Mz_2 \right) ^{q_c}\\
			=&- \sum_{i=1}^n{\frac{z_{1_i}z_{2_i}k_{c_i}^{2}}{k_{c_i}^{2}-{\eta _{1_i}}^2}}-\lambda _3V_2  ^{p_c}-\lambda _4V_2 ^{q_c}
		\end{aligned} 
	\end{equation}
\end{small}
where
	\begin{equation}
		\label{s3_17}
		\begin{aligned}
			&\lambda _3=\frac{\lambda _{\min}\left( k_2 \right)}{\lambda _{\max}\left( M \right) ^{p_c}} n^{1-{p_c}}
			\\
			&\lambda _4=\frac{\lambda _{\min}\left( k_3 \right)}{\lambda _{\max}\left( M \right) ^{q_c}}
		\end{aligned} 
	\end{equation}

Therefore, according to Lemma 4, the derivative of the Lyapunov function for the proposed controller is:
\begin{equation}
	\label{s3_18}
	\begin{aligned}
		\dot{V}=\dot{V}_1+\dot{V}_2&\le-\lambda _1V_{1}^{p_c}-\lambda _2V_{1}^{{q_c}}-\lambda _3V_{2}^{p_c}-\lambda _4V_{2}^{{q_c}}\\
		&\le -v_1V^{p_c}-v_2V^{q_c}
	\end{aligned} 
\end{equation}
where

\begin{equation}
	\label{s3_19}
	\begin{aligned}
		&v_1=2^{1-{p_c}}\min \left( \lambda _1,\lambda _3 \right) 
		\\
		&v_2=\min \left( \lambda _2,\lambda _4 \right) 
	\end{aligned} 
\end{equation}

According to Lemma 1, the proposed controller $u$ is fixed-time stable. Assuming the robot dynamics are unknown. In the controller $u$ defined in (\ref{s3_14}), we collect the terms that contains dynamics and disturbances into a function $D\left( Z \right) \in R^n$, and then design NNs to approximate it, that is: 
\begin{equation}
	\label{s3_20}
	\begin{aligned}
		D\left( Z \right) =-G_x-F_x-M_x\dot{\alpha}-C_x\alpha={W}^TS\left( Z \right) +\epsilon 
	\end{aligned} 
\end{equation}
Each joint of the robot manipulator is assigned it own NN for approximation, hence, $D\left( Z \right) =\left[ D_1\left( Z_1 \right) ,D_2\left( Z_2 \right) ,...D_n\left( Z_n \right) \right] ^T$. The $i$-th element of $D\left( Z \right)$ is given by:
\begin{equation}
	\label{s3_21}
	\begin{aligned}
		D_i\left( Z_i \right) ={W_i}^TS_i\left( Z_i \right) +\epsilon _i 
	\end{aligned} 
\end{equation}
where $Z_i=\left[z_{i_1},z_{i_2},...,z_{i_r} \right] ^T$, $W_i=\left[ w_{i_1},w_{i_2},...,w_{i_l} \right]^T $, and $S_i\left( Z_i \right) =\left[ s_{i_1}\left( Z_i \right) ,s_{i_2}\left( Z_i \right) ,...,s_{i_l}\left( Z_i \right) \right] ^T$ are the input vector, weight vector and hidden layer output of the $i$-th NN, respectively. Each hidden node comprises a Gaussian radial basis function $s_{i_l}$ as defined in (\ref{RBFNN1}), and $Z_i=\left[q_1,...,q_n,\dot{q}_1,...,\dot{q}_n,\alpha_1,...,\alpha_n,\dot{\alpha_1},...,\dot{\alpha_n}\right]^T$ is the vector of input variables. To enable a compact matrix representation, in (\ref{s3_20}), $W\in R^{n\times nl}$ is defined as
$W=\mathrm{diag}\left[ W_{1}^{T},W_{2}^{T},...,W_{n}^{T} \right]$ and $S\left(Z\right) \in R^{nl \times 1}$ is defined as $S=\left[ S_1,S_2,...,S_n \right] ^T$. $\epsilon=\left[\epsilon_1,\epsilon_2,...,\epsilon_n\right]^T$ is the vector of the estimation errors of each joint.

Denoting $\hat{W_i}$ as the estimate of $W_i$, and the estimation error as  $\tilde{W_i}=W_i-\hat{W_i}$, we design the adaptive NN update law as:

\begin{equation}
	\label{s3_22}
	\begin{aligned}
		\dot{\hat{W_i}}=S_i\left( Z_i \right) z_{2_i}-k_4\hat{W_i}^{2p_c-1}-k_5\hat{W_i}^{2q_c-1}
	\end{aligned} 
\end{equation}

\noindent By using NNs, the formulation of controller $u$ becomes:

\begin{small}
	\begin{equation}
		\label{s3_32}
		\begin{aligned}
			u=&-\hat{W}^TS\left( Z \right) -f_e-\frac{k_{c}^{2}z_1}{k_{c}^{2}-\eta _{1}^{2}}-k_1z_2\\
			&-\frac{1}{2^{p_c}}k_2z^{2p_c-1}-\frac{1}{2^{q_c}}k_3z^{2q_c-1}
		\end{aligned} 
	\end{equation}
\end{small}

\noindent\textbf{\emph{Remark 2}}: The differences between the proposed NN adaptive law and the traditional adaptive laws \cite{RN103} \cite{RN584} are that the extra terms added at the end of the formula can ensure the overall fixed-time stability of the control system.

\noindent\textbf{\emph{Remark 3}}: When implementing the proposed adaptive law (\ref{s3_22}), to avoid the singularity problem when $\hat{W}_i<0$, we can replace the $-k_4\hat{W_i}^{2p_c-1}$ term with  $-k_4sign\left( \hat{W}_i \right) \left| \hat{W}_i \right|^{2p_c-1}$ (similar to the controller $u$).

\noindent\textbf{\emph{Theorem 2}}: Consider the system (\ref{eq1}) subject to the constraints $k_c\left(t\right)$, external disturbances $F\left(q, \dot{q} \right)$ and unknown dynamics $M\left( q \right)$, $C\left( q,\dot{q} \right)$ and $G\left( q \right)$, the system is overall fixed-time stable and the convergence time of the tracking errors are bounded by applying the stabilizing function (\ref{s3_9}), controller (\ref{s3_32}) and the NNs adaptive law (\ref{s3_22}).

\noindent\textbf{\emph{Proof}}: Select a Lyapunov function candidate $V_3$ as:
\begin{equation}
	\label{s3_23}
	\begin{aligned}
		V_3=\frac{1}{2}\sum_{i=1}^n{{\tilde{W}_i}^T\tilde{W}_i}
	\end{aligned} 
\end{equation}

The derivative of $V_3$ is:
\begin{small}
	\begin{equation}
		\label{s3_24}
		\begin{aligned}
			\dot{V}_3&=-\sum_{i=1}^n{{\tilde{W}_i}^T\dot{\hat{W}}_{i}}
			\\
			&=\sum_{i=1}^n{{\tilde{W}_{i}}^T\left(- S_{i}\left( Z \right) z_{2_i}+k_4{\hat{W}_{i}}^{2p_c-1}+k_5{\hat{W}_{i}}^{2q_c-1} \right)}
			\\
			&=\sum_{i=1}^n{-{\tilde{W}_{i}}^TS_{i}\left( Z \right) z_{2_i}+}k_4\sum_{i=1}^n{{\tilde{W}_{i}}^T{\hat{W}_{i}}^{2p_c-1}+k_5\sum_{i=1}^n{{\tilde{W}_{i}}^T{\hat{W}_{i}}^{2q_c-1}}}
		\end{aligned} 
	\end{equation}
\end{small}

According to Lemma 3, we have:
\begin{equation}
	\label{s3_25}
	\begin{aligned}
		{\tilde{W}_{i}}{\hat{W}_{i}}^{2q_c-1}\le n_1W_{i}^{2q_c}-n_2\tilde{W}_{i}^{2q_c}
	\end{aligned} 
\end{equation}

Furthermore, according to Lemma 6, $\hat{W}\left( t \right) \ge 0 $ is true if $\hat{W}\left( t_0 \right) \ge 0 $. Since $\tilde{W}=W-\hat{W}$, it is clear that $\tilde{W}\le W$. Therefore, according to Lemma 5, we have:
\begin{equation}
	\label{s3_26}
	\begin{aligned}
		\tilde{W}_{i}{\hat{W}_{i}}^{2p_c-1}&=\tilde{W}_{i}\left( W_i-\tilde{W}_{i} \right) ^{2p_c-1}
		\\
		&\le W_i^{2p_c}-\tilde{W_i}^{2p_c}
	\end{aligned} 
\end{equation}

Applying Lemma 4, we obtain:
\begin{small}
	\begin{equation}
		\label{s3_27}
		\begin{aligned}
			\dot{V}_3\le& -\sum_{i=1}^n{{\tilde{W}_i}^TS_i\left( Z_i \right) z_{2_i}+}k_4\sum_{i=1}^n{\sum_{j=1}^l{\left( {w_{ij}}^{2p_c}-{\tilde{w}_{ij}}^{2p_c} \right)}}\\
			&+k_5\sum_{i=1}^n{\sum_{j=1}^l{\left( n_1{w_{ij}}^{2q_c}-n_2{\tilde{w}_{ij}}^{2q_c} \right)}}
			\\
			=&-\sum_{i=1}^n{{\tilde{W}_i}^TS_i\left( Z_i \right) z_{2_i}}-k_4\sum_{i=1}^n{\sum_{j=1}^l{{\tilde{w}_{ij}}^{2p_c}}}\\
			&-k_5n_2\sum_{i=1}^n{\sum_{j=1}^l{{\tilde{w}_{ij}}^{2q_c}}}+\sigma 
		\end{aligned} 
	\end{equation}
\end{small}
where

\begin{small}
	\begin{equation}
		\label{s3_28}
		\begin{aligned}
			\sigma =k_4\sum_{i=1}^n{\sum_{j=1}^l{{w_{ij}}^{2p_c}}}+k_5n_1\sum_{i=1}^n{\sum_{j=1}^l{{w_{ij}}^{2q_c}}}
		\end{aligned} 
	\end{equation}
\end{small}

\noindent Again, using Lemma 4, we have:
\begin{small}
	\begin{equation}
		\label{s3_29}
		\begin{aligned}
			\dot{V}_3\le& -\sum_{i=1}^n{{\tilde{W}_i}^TS_i\left( Z_i \right) z_{2_i}}-k_4l^{1-{p_c}}\sum_{i=1}^n{\left( \sum_{j=1}^l{{\tilde{w}_{ij}}^2} \right) ^{p_c}}\\
			&-k_5n_2\sum_{i=1}^n{\left( \sum_{j=1}^l{{\tilde{w}_{ij}}^2} \right) ^{q_c}}+\sigma
			\\
			=&-\sum_{i=1}^n{{\tilde{W}_i}^TS_i\left( Z_i \right) z_{2_i}}-2^p_c k_4l^{1-{p_c}}\frac{1}{2^{p_c}}\sum_{i=1}^n{\left( {\tilde{W}_i}^T\tilde{W}_i \right) ^{p_c}}\\
			&-2^q_c k_5n_2\frac{1}{2^{q_c}}\sum_{i=1}^n{\left( {\tilde{W}_i}^T\tilde{W}_i \right) ^{q_c}}+\sigma
			\\
			\le& -\sum_{i=1}^n{{\tilde{W}_i}^TS_i\left( Z_i \right) z_{2_i}}-\lambda _5\left( \frac{1}{2}\sum_{i=1}^n{{\tilde{W}_i}^T\tilde{W}_i} \right) ^{p_c}\\
			&-\lambda _6\left( \frac{1}{2}\sum_{i=1}^n{{\tilde{W}_i}^T\tilde{W}_i} \right) ^{q_c}+\sigma
		\end{aligned} 
	\end{equation}
\end{small}

\noindent Therefore, it follows that:
\begin{equation}
	\label{s3_30}
	\begin{aligned}
		\dot{V}_3\le -\sum_{i=1}^n{{\tilde{W}_i}^TS_i\left( Z_i \right) z_{2_i}}-\lambda _5 V_3  ^{p_c}-\lambda _6V_3  ^{q_c}+\sigma
	\end{aligned} 
\end{equation}
where 

\begin{equation}
	\label{s3_31}
	\begin{aligned}
		&\lambda _5=2^{p_c}k_4l^{1-{p_c}}n^{1-{p_c}}
		\\
		&\lambda _6=2^{q_c}k_5n_2
	\end{aligned} 
\end{equation}

\noindent Inserting the controller (\ref{s3_32}) into $\dot{V}_2$ gives:
\begin{small}
	\begin{equation}
		\label{s3_33}
		\begin{aligned}
			\dot{V}_2\le z_{2}^{T}\tilde{W}^TS\left( Z \right) -\sum_{i=1}^n{\frac{z_{1_i}z_{2_i}k_{c_i}^{2}}{k_{c_i}^{2}-\eta _{1_i}^{2}}}-\lambda _3 V_2  ^{p_c}-\lambda _4 V_2  ^{q_c}-z_{2}^{T}\epsilon 
		\end{aligned} 
	\end{equation}
\end{small}

\noindent According to Lemma 7 and Assumption 1:
\begin{equation}
	\begin{aligned}
		\label{s3_35}
		-z_{2}^{T}\epsilon \le \frac{1}{2}z_{2}^{T}z_2+\frac{1}{2}\left\| \overline{\epsilon } \right\| ^2 
	\end{aligned}
\end{equation}

\noindent Therefore:
\begin{small}
	\begin{equation}
		\label{s3_36}
		\begin{aligned}
			\dot{V}_2\le& z_{2}^{T}\tilde{W}^TS\left( Z \right) -\sum_{i=1}^n{\frac{z_{1_i}z_{2_i}k_{c_i}^{2}}{k_{c_i}^{2}-\eta _{1_i}^{2}}}-z_{2}^{T}\left( k_1-\frac{1}{2}I \right) z_2\\
			&-\lambda _3 V_2  ^{p_c}-\lambda _4 V_2  ^{q_c}+\frac{1}{2}\left\| \bar{\epsilon} \right\| ^2
		\end{aligned} 
	\end{equation}
\end{small}

\noindent where $k_1$ is a parameter matrix to be designed such that $k_1-\frac{1}{2}I$ is positive. Letting $V=V_1+V_2+V_3$, according to (\ref{s3_11}), (\ref{s3_36}) and (\ref{s3_30}), we can write:

\begin{small}
\begin{equation}
	\begin{aligned}
		\label{s3_37}
		\dot{V}&\le -\lambda _1V_{1}^{p_c}-\lambda _2V_{1}^{{q_c}}-\lambda _3{V_2}^{p_c}-\lambda _4{V_2}^{q_c}-\lambda _5V_{3}^{{p_c}}-\lambda _6V_{3}^{{q_c}} \\
		&\le -\alpha V^{p_c}-\beta V^{q_c} + \sigma
	\end{aligned}
\end{equation}
\end{small}
where

\begin{small}
	\begin{equation}
		\begin{aligned}
			\label{s3_38}
			&\alpha =3^{1-{p_c}}\min \left( \lambda _1,\lambda _3,\lambda _5 \right) 
			\\
			&\beta =\min \left( \lambda _2,\lambda _4,\lambda _6 \right) 
		\end{aligned}
	\end{equation}
\end{small}

\noindent and therefore by employing Lemma 1, the proposed controller combined with NNs is fixed-time stable, and that the convergence time of its tracking error is bounded by:
\begin{equation}
	T\le T_{\max}:=\frac{1}{\alpha v \left( {p_c}-1 \right)}+\frac{1}{\beta v \left( 1-{q_c} \right)} 
\end{equation}

\section{Simulation Example}\label{sec4}
\begin{figure}[h]
	\centering
	\includegraphics[width=0.32\textwidth]{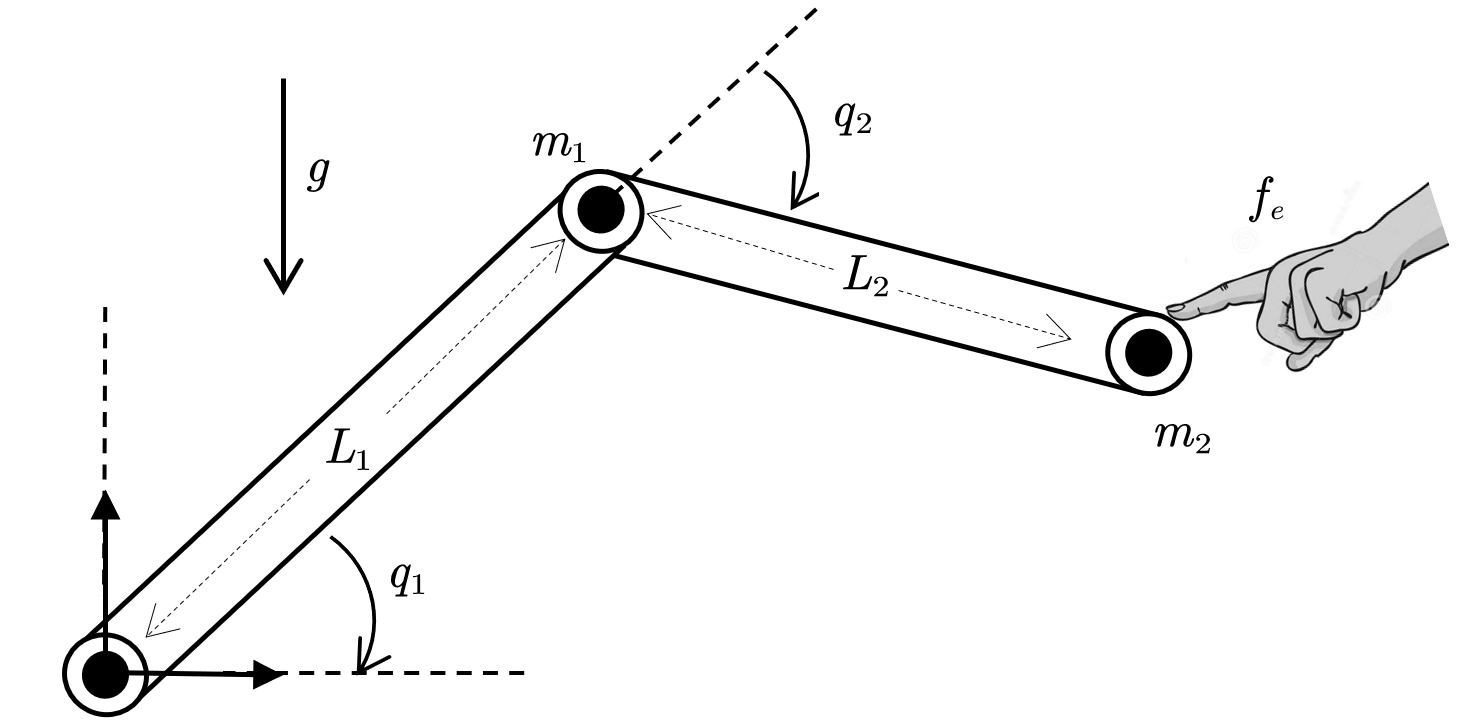}
	\caption{A two-link planar robot manipulator.}
	\label{Twolink}
\end{figure}

In this section, a comparative simulation based on a two-link planar robot manipulator is employed to show the performance of the proposed method, as depicted in Fig. \ref{Twolink}. The proposed method can in principle be extended to robots with arbitrary degrees of freedom.

The dynamics of the robot are given by \cite{RN1987}:
\begin{small}
	\begin{equation*}
		\begin{aligned}
			\tau _1&=m_2l_{2}^{2}\left( \ddot{q}_1+\ddot{q}_2 \right) +m_2l_1l_2c_2\left( 2\ddot{q}_1+\ddot{q}_2 \right) +\left( m_1+m_2 \right) l_{1}^{2}\ddot{q}_1-\\
			&m_2l_1l_2s_2\dot{q}_{2}^{2}-2m_2l_1l_2s_2\dot{q}_1\dot{q}_2+m_2l_2gc_{12}+\left( m_1+m_2 \right) l_1gc_1\\
			\tau _2&=m_2l_1l_2c_2\ddot{q}_1+m_2l_1l_2s_2\dot{q}_{1}^{2}+m_2l_2gc_{12}+m_2l_{2}^{2}\left( \ddot{q}_1+\ddot{q}_2 \right) 
		\end{aligned}
	\end{equation*}
\end{small}

\noindent where $c_i=\cos \left( q_i \right) $, $c_{ij}=\cos \left( q_i+q_j \right) $, $s_i=\sin \left( q_i \right) $, and $s_{ij}=\sin \left( q_i+q_j \right)$, $i,j=1,2$. The coefficient matrices $M\left( q \right)$, $C\left( q,\dot{q} \right)$, and $G\left( q \right)$ are given as: 
\begin{small}
	\begin{equation*}
		\begin{split}
			M\left( q \right) =\left[ \begin{matrix}
				m_2l_{2}^{2}+2m_2l_1l_2c_2+\left( m_1+m_2 \right) l_{1}^{2}&		m_2l_{2}^{2}+m_2l_1l_2c_2\\
				m_2l_{2}^{2}+m_2l_1l_2c_2&		m_2l_{2}^{2}\\
			\end{matrix} \right] \ \ \ \ \ \ \ \ \ \ \ \ \ \ \ \ \\
			C\left( q,\dot{q} \right) =\left[ \begin{matrix}
				-2m_2l_1l_2s_2\dot{q}_2&		-m_2l_1l_2s_2\\
				m_2l_1l_2s_2\dot{q}_1&		0\\
			\end{matrix} \right] \ \ \ \ \ \ \ \ \ \ \ \ \ \ \ \ \ \ \ \ \ \ \ \ \ \ \ \ \ \ \ \ \ \\
			G\left( q \right) =\left[ \begin{array}{c}
				m_2l_2gc_{12}+\left( m_1+m_2 \right) l_1gc_1\\
				m_2l_2gc_{12}\\
			\end{array} \right]\ \ \ \ \ \ \ \ \ \ \ \ \ \ \ \ \ \ \ \ \ \ \ \ \  \ \ \ \ \ \ \\
		\end{split}
	\end{equation*}
\end{small}

\noindent and the uncertainties and disturbances term is defined as:
\begin{small}
	\begin{equation*}
		\begin{split}
			F\left( q,\dot{q} \right) =\left[ \begin{array}{c}
				4c_1s_2 + 6c_{1}^{2}-2\\
				-4c_1s_2-6c_{1}^{2}+2\\
			\end{array} \right]
		\end{split}
	\end{equation*}
\end{small}

The Jacobian of the robot is given by:
\begin{small}
	\begin{equation*}
		\begin{split}
			J\left( q \right) =\left[ \begin{matrix}
				-l_1s_1-l_2s_{12}&		-l_2s_{12}\\
				l_1c_1+l_2c_{12}&		l_2c_{12}\\
			\end{matrix} \right]
		\end{split}
	\end{equation*}
\end{small}

\begin{figure}[h]
	\centering
	\includegraphics[width=0.4\textwidth]{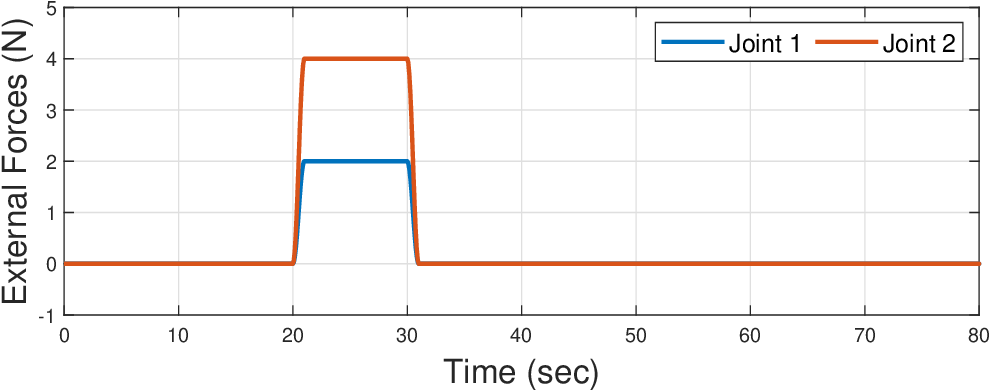}
	\caption{External human forces.}
	\label{fe_1}
\end{figure}

External human forces $f_{e_i}$ are applied to each link of the robot at 20s and removed at 31s with profiles as specified in (\ref{fe}).  Here, $a=\left[a_1, a_2\right]$ are link specific force scaling parameters. Fig. \ref{fe_1} shows the evolution of the applied forces over time.
\begin{small}
	\begin{equation}
		\label{fe}
		f_{e_i}\left( t \right) =\left\{ \begin{array}{l}
			0\ \ \ \ \ \ \ \ \ \ \ \ \ \ \ \ \ \ \ \ t<20\ or\ t\ge 31\\
			a_i\left( 1-\cos \pi t \right) \ \ \ \ \ \ 20\le t<21\\
			2a_i\ \ \ \ \ \ \ \ \ \ \ \ \ \ \ \ \ \ \ \  21\le t<30\\
			a_i\left( 1+\cos \pi t \right) \ \ \ \ \ \ 30\le t<31
		\end{array} \right. 
	\end{equation}
\end{small}

The designed trajectory is given as:
\begin{equation}
		\label{xd}
		\begin{aligned}
			x_{d_1}\left( t \right) &= 0.18\cos \left( 0.5t \right) \\
			x_{d_2}\left( t \right) &=0.18\sin \left( 0.5t \right) 	
		\end{aligned}
\end{equation}

\begin{figure}[h]
	\centering
	\includegraphics[width=0.4\textwidth]{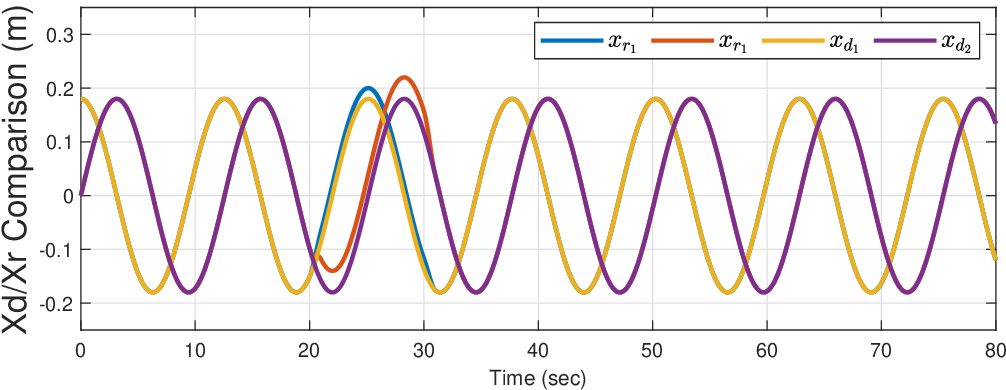}
	\caption{Comparison of desired and reference trajectory.}
	\label{xr_xd}
\end{figure}

When we have $x_{d_i}$ and $f_{e_i}$, the reference trajectory which can comply with human forces is derived by integrating (\ref{eq10}) twice, as depicted in Fig. \ref{xr_xd}. The $x_d$ and $x_r$ coincide when $f_e=0$. When $20<t<31$, the desired trajectory is modified by the external forces to comply with human forces.  

The time-varying workspace safety constraints are given as:

\begin{equation}
		\label{kc}
		\begin{aligned}
			k_{c_1}&=0.48+0.1\cos \left( 0.2t-\frac{\pi}{3} \right) 
			\\
			k_{c_2}&=-0.48+0.1\sin \left( 0.2t \right) 
		\end{aligned}
\end{equation}

\begin{table}[t]
	\centering
	\caption{Simulation parameters}
	\renewcommand\arraystretch{2.5}
	\begin{tabular}{cc}  
		\hline
		\hline
		Modules & Parameters\\  
		\hline
		Initial values &  \makecell{$q\left(0\right)=\left[0.5236,2.0944\right]^T$ \\ $x\left(0\right)=\left[0,0\right]^T$ } \\ 
		\hline
		Robot dynamics & \makecell{$m_1=1.5kg$, $m_2=1.0kg$ \\$l_1=l_2=0.3m$} \\ 
		\hline
		Admittance control &   \makecell{$k_{m_i}=20$, $k_{b_i}=20$, $k_{k_i}=100$ \\ $a_1=1$, $a_2=2$} \\ 
		\hline
		Controller &  \makecell{${p_c}=3$, ${q_c}=\frac{99}{101}$ \\ $\theta_1=\left[10,0.01\right]$, $\theta_2=\left[20,0.01\right]$ ,$\kappa_1=\left[5,22\right]$\\
		$k_1=diag\left[5,22\right]$, $k_2=diag\left[100,2000\right]$\\ 
		$k_3=diag\left[200,3000\right]$, $k_4=k_5=0.001$}\\ 
		\hline
		NNs &  \makecell{$l=8$, $Z=\left[q_1,q_2,\dot{q}_1,\dot{q}_2,\alpha_1,\alpha_2,\dot{\alpha_1},\dot{\alpha_2}\right]^T$\\
		$C=\left[-25,-15,-5,-1,1,5,15,25\right]$, $B=40$}  \\ 
		\hline 
		\hline 
	\end{tabular}
\end{table}

To verify the performance of the proposed controller FxTTVIBLF, we compare it with IBLF \cite{RN103} and traditional TVIBLF based controllers without fixed-time terms \cite{RN588}. To further illustrate the effectiveness of the proposed NNs, we divide the simulation into two cases: model-based and model-free. 

\noindent\textbf{\emph{Remark 4}}: Since there is no existing literature on TVIBLF published in the area of robot manipulators, to illustrate the improvement of the proposed fixed-time controller, the TVIBLF we applied in the comparative simulation is a simplified version of the controller proposed in this paper. That is, we delete the fixed-time terms in the proposed controller. For details of TVIBLF controller design see \cite{RN588}.

\subsection{Model-based control}
\begin{figure}[htbp]
	\centering
	\begin{subfigure}{0.99\linewidth}
		\centering
		\includegraphics[width=0.9\linewidth]{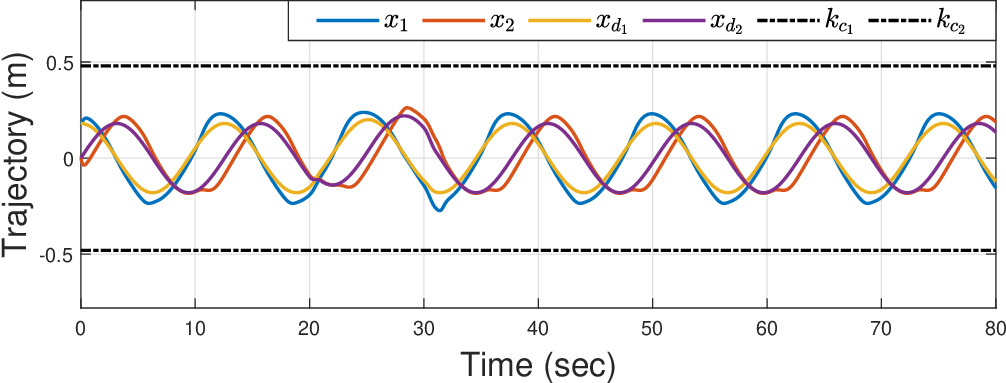}
		\caption{IBLF controller}
		\label{f1_1}
	\end{subfigure}
	\centering
	\begin{subfigure}{0.99\linewidth}
		\centering
		\includegraphics[width=0.9\linewidth]{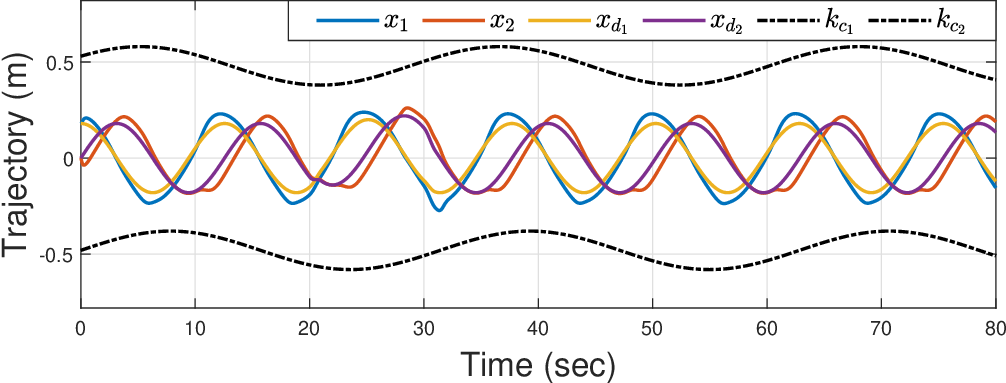}
		\caption{TVIBLF controller}
		\label{f1_2}
	\end{subfigure}
	\centering
	\begin{subfigure}{0.99\linewidth}
		\centering
		\includegraphics[width=0.9\linewidth]{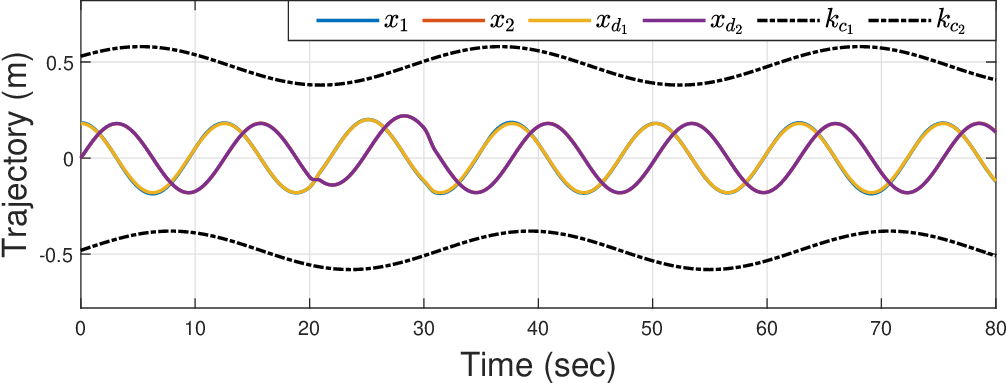}
		\caption{FxTTVIBLF controller }
		\label{f1_3}
	\end{subfigure}
	\caption{Trajectories of compared controllers without NNs}
	\label{f1}
\end{figure}

First, we assume all of the robot dynamics are known except for disturbances term $F\left(q,\dot{q}\right)$. The parameters of the robot dynamics and proposed controller are given in Table I. Fig. \ref{f1} shows the task space trajectories of the compared controllers. We can see that the proposed FxTTVIBLF can follow the reference $x_r$ within the time-varying constraint bounds.  Moreover, the tracking accuracy of the proposed controller is better than achieved with IBLF and TVIBLF based controllers, which means the proposed controller has the best immunity to uncertainties and tracking performance is improved by integrating fixed-time terms. The RMSE tracking performance of each controller is shown in Table II.

\subsection{Model-free control}
\begin{figure}[htbp]
	\centering
	\begin{subfigure}{0.99\linewidth}
		\centering
		\includegraphics[width=0.9\linewidth]{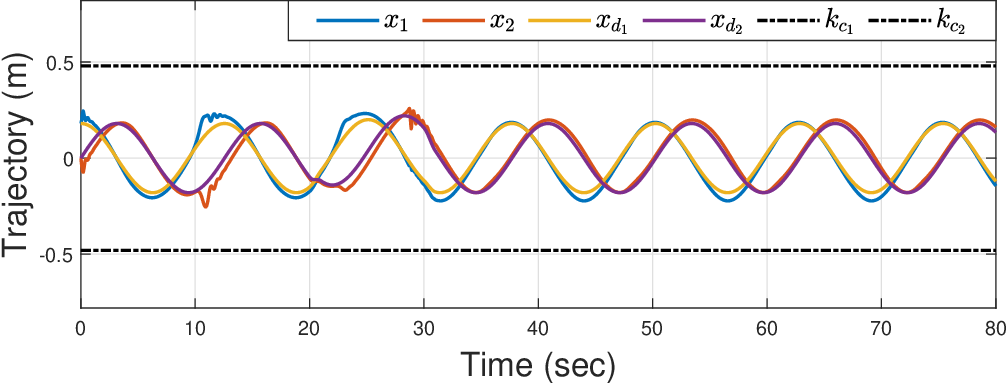}
		\caption{IBLF controller}
		\label{f2_1}
	\end{subfigure}
	\centering
	\begin{subfigure}{0.99\linewidth}
		\centering
		\includegraphics[width=0.9\linewidth]{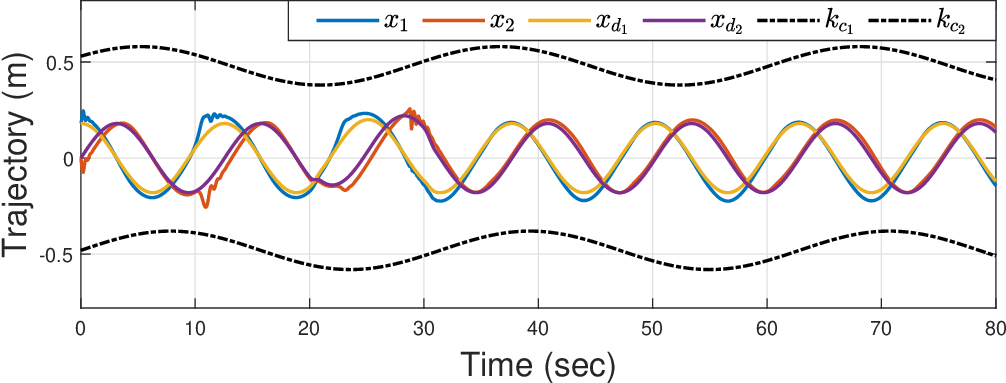}
		\caption{TVIBLF controller}
		\label{f2_2}
	\end{subfigure}
	\centering
	\begin{subfigure}{0.99\linewidth}
		\centering
		\includegraphics[width=0.9\linewidth]{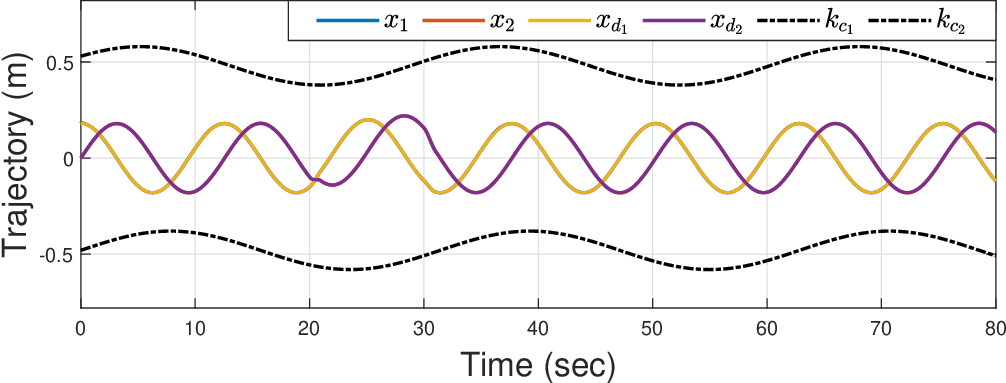}
		\caption{FxTTVIBLF controller}
		\label{f2_3}
	\end{subfigure}
	\caption{Trajectories of compared controllers with NNs}
	\label{f2}
\end{figure}

Here we assume that all the parameters of the robot dynamics are unknown. We employ an NN (with traditional adaptive law \cite{RN103}) with IBLF and TVIBLF for the contrast simulation, and employ an NN (with our proposed adaptive law (\ref{s3_22})) with FxTTVIBLF (\ref{s3_32}) to achieve overall fixed-time stability. By doing so, the three controllers are capable of estimating the uncertainties along with the dynamics. Fig. \ref{f2} shows the tracking trajectories of the controllers. We can see the performance of all these controllers is improved to some extent. Notably, the trajectory of the proposed FxTTVIBLF controller coincides with $x_r$ much more quickly that observed with the other controllers, which is a consequence of the overall faster converge properties of the fixed-time control law combining with the proposed NNs adaptive law.

\begin{figure}[htbp]
	\centering
	\begin{subfigure}{0.99\linewidth}
		\centering
		\includegraphics[width=0.9\linewidth]{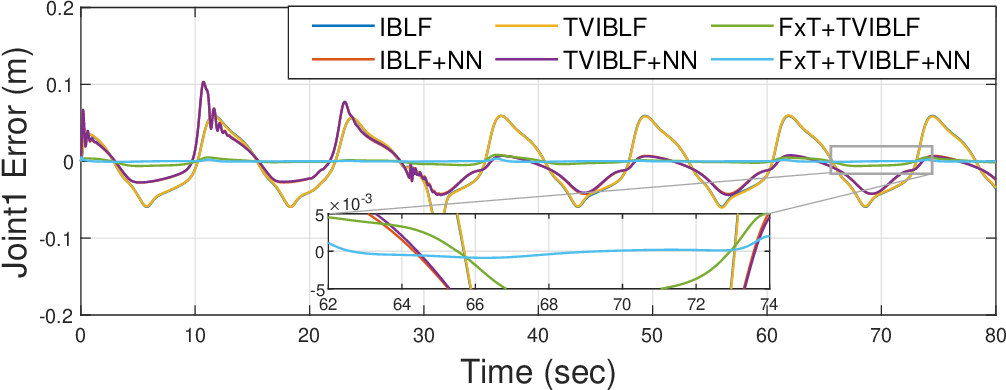}
		\caption{Joint 1}
		\label{f3_1}
	\end{subfigure}
	\centering
	\begin{subfigure}{0.99\linewidth}
		\centering
		\includegraphics[width=0.9\linewidth]{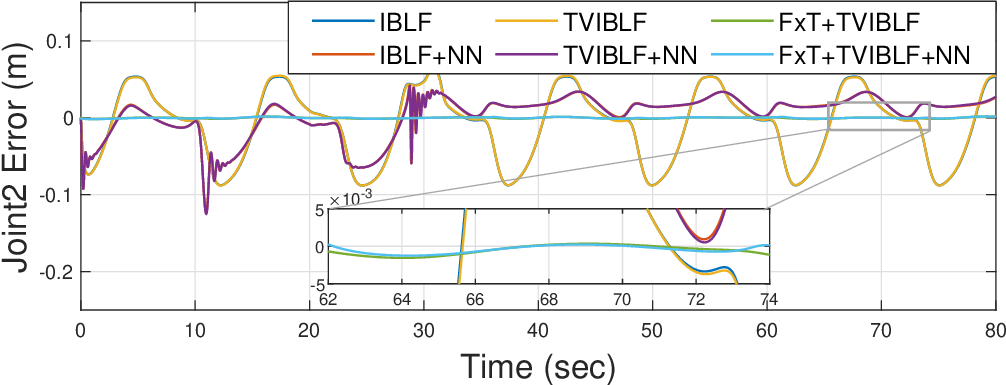}
		\caption{Joint 2}
		\label{f3_2}
	\end{subfigure}
	\caption{Tracking errors of compared controllers}
	\label{f3}
\end{figure}

\begin{figure}[htbp]
	\centering
	\begin{subfigure}{0.99\linewidth}
		\centering
		\includegraphics[width=0.9\linewidth]{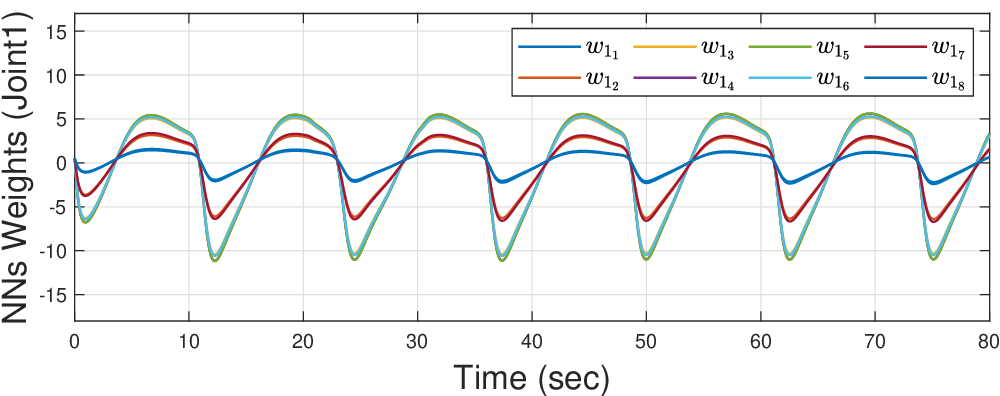}
		\caption{Joint 1}
		\label{NN_1}
	\end{subfigure}
	\centering
	\begin{subfigure}{0.99\linewidth}
		\centering
		\includegraphics[width=0.9\linewidth]{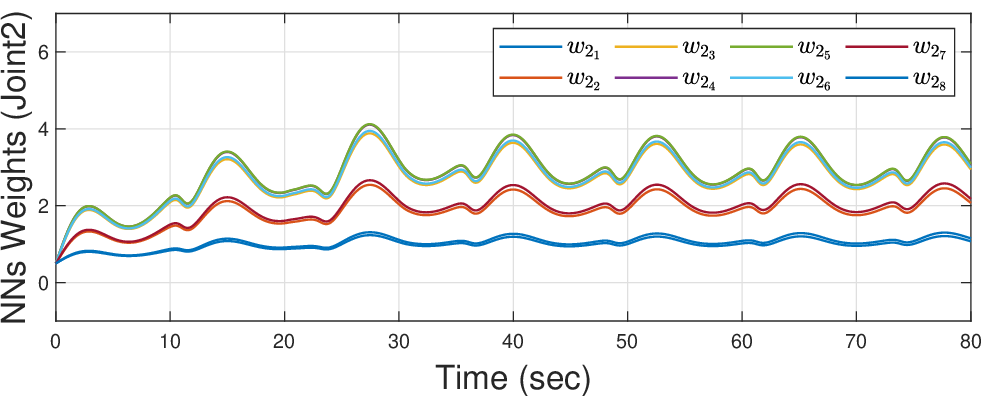}
		\caption{Joint 2}
		\label{NN_2}
	\end{subfigure}
	\caption{NNs weights evolution of FxTTVIBLF+NN}
	\label{NN_output}
\end{figure}

\begin{figure}[htbp]
	\centering
	\begin{subfigure}{0.99\linewidth}
		\centering
		\includegraphics[width=0.9\linewidth]{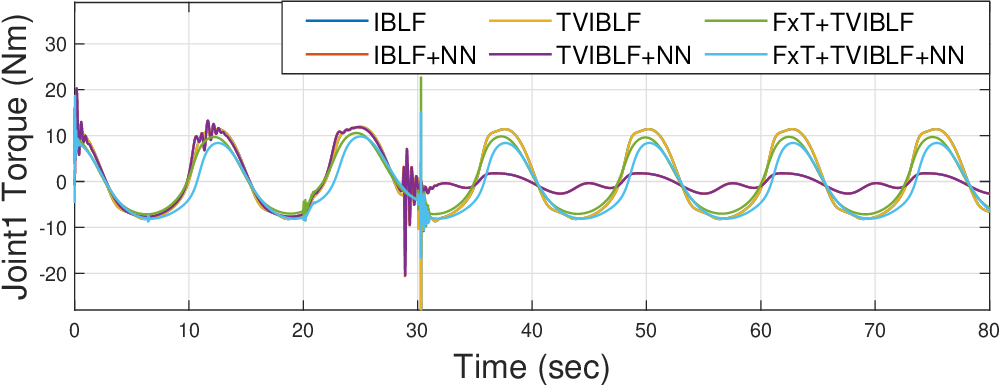}
		\caption{Joint 1}
		\label{f4_1}
	\end{subfigure}
	\centering
	\begin{subfigure}{0.99\linewidth}
		\centering
		\includegraphics[width=0.9\linewidth]{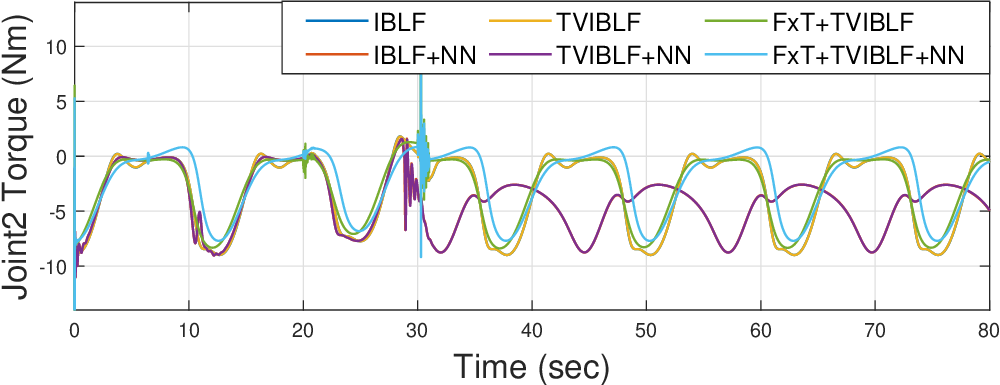}
		\caption{Joint 2}
		\label{f4_2}
	\end{subfigure}
	\caption{Control effort of compared controllers}
	\label{f4}
\end{figure}

Fig. \ref{f3} shows the tracking errors of the various model-based and model-free controllers considered. The errors of the IBLF and TVIBLF based controllers combined with traditional NNs show chattering at the beginning and gradually converge after 40s. It is evident that the proposed controller achieves a smoother trajectory, smaller tracking error and overall faster convergence time than the other controllers.
Fig. \ref{NN_output} shows the evolution of the NN weights vectors of the proposed FxTTVIBLF. All weights of the hidden nodes are initialized as zero at 0s and updated in real time. Table II compares the tracking RMSE performance of each controller. It is clear that the tracking performance of IBLF and TVIBLF is similar. The tracking error is smaller when we apply NNs and integrate fixed-time techniques into the control design. Fig. \ref{f4} shows the control effort of each controller. It can be seen that the proposed controller generally requires a similar control effort despite the demonstrated performance advantages over the other controllers. However, it is noted that some chattering occurs with FxTTVIBLFs when the external forces are applied and removed (at 20s and 30s).
\begin{center}
	\label{table1}
	\begin{table}[t]
		\caption{RMSE tracking performance of each controller}
		\setlength{\tabcolsep}{3mm}{
			\begin{tabular}{ccccccc}
				\toprule
				\toprule
				\makecell[c]{Joint} &\makecell[c]{IBLF}& \makecell[c] {TVIBLF}  &\makecell[c]{FxTTVIBLF}\\				
				\midrule
				\multirowcell{0.1}{1}&$3.74\times 10^{-2}$&$3.73\times 10^{-2}$&$3.2\times 10^{-3}$\\
				\midrule
				\multirowcell{0.1}{2}&$4.68\times 10^{-2}$&$4.69\times 10^{-2}$&$8.63\times 10^{-4}$\\	
				\midrule
				\midrule
				\makecell[c]{Joint} &\makecell[c]{IBLF+NN}& \makecell[c] {TVIBLF+NN}  &\makecell[c]{FxTTVIBLF+NN}\\				
				\midrule
				\multirowcell{0.1}{1}&$2.56\times 10^{-2}$&$2.56\times 10^{-2}$&$6.10\times 10^{-4}$\\
				\midrule
				\multirowcell{0.1}{2}&$2.73.59\times 10^{-2}$&$2.74\times 10^{-2}$&$7.71\times 10^{-4}$\\		
				\bottomrule
				\bottomrule
		\end{tabular}}
	\end{table}
\end{center}

\section{Conclusions}\label{sec5}
In this paper, a fixed-time time-varying IBLF controller based on admittance control has been proposed for physical human-robot collaboration. The proposed approach guarantees both safety and compliance during physical contact. Compared with existing methods, the proposed controller has lower tracking error, faster convergence time and more human-friendly behaviour which makes it more practical in real-life scenarios. The BLF based constraint control strictly guarantees that the resultant trajectory never violates the preset bounds. When the desired trajectory traverses beyond these bounds the robot will stop because the nature of the BLF. In the future we will build on this work and explore its use with a high-level path planning block to allow the system to achieve real time dynamic obstacle avoidance. In this context the proposed FxTTVIBLF controller will act as the low level controller to strictly guarantee safety.





\bibliographystyle{IEEEtran}
\bibliography{References}

\end{document}